\documentclass[preprint,12pt]{elsarticle}

\usepackage{amssymb}
\usepackage{amsmath}
\usepackage{booktabs}
\usepackage{multirow}
\usepackage{array}
\usepackage{float}
\usepackage{url}
\usepackage{placeins}
\usepackage[ruled,vlined]{algorithm2e}

\usepackage{colortbl}
\usepackage[table]{xcolor}

\definecolor{best}{RGB}{198,239,206}   % green
\definecolor{second}{RGB}{226,239,218} % light green
\definecolor{third}{RGB}{255,242,204}  % yellow

\journal{Nuclear Physics B}

\begin{document}

\begin{frontmatter}



% \title{Fast SAM2 with Text-Driven Token Pruning}


%% Title
\title{Fast SAM2 with Text-Driven Token Pruning}

%% Authors
\author[1,2]{Avilasha Mandal}
%\ead{avilasha1112@gmail.com, cs1221631@iitd.ac.in}

\author[1]{Chaoning Zhang\corref{cor1}}
\ead{chaoningzhang1990@gmail.com}
\cortext[cor1]{Corresponding author}

\author[1]{ Fachrina Dewi Puspitasari}
%\ead{third.author@institute-b.org}

\author[1]{Xudong Wang}
%\ead{fourth.author@company-c.com}

\author[1]{ Jiaquan Zhang}
%\ead{jiaquanzhang2005@gmail.com}

\author[3]{Caiyan Qin}
%\ead{qincaiyan@hit.edu.cn}
%% Corresponding author text

\author[1]{Guoqing Wang}

\author[1]{Yang Yang}

\author[1]{Heng Tao Shen}
%% Affiliations

\affiliation[1]{
  organization={School of Computer Science and Engineering, University of Electronic Science and Technology of China},
  %addressline={},
  city={Chengdu},
  postcode={610054},
  state={Sichuan},
  country={China}
}

\affiliation[2]{
  organization={Department of Computer Science and Engineering, Indian Institute of Technology, Delhi},
  %addressline={Hauz Khas},
  city={New Delhi},
  postcode={110016},
  state={Delhi},
  country={India}
}

\affiliation[3]{
  organization={School of Robotics and Advanced Manufacture, Harbin Institute of Technology},
  %addressline={Shenzhen},
  city={Shenzhen},
  postcode={518055},
  state={Guangdong},
  country={China}
}

\begin{abstract}
Segment Anything Model 2 (SAM2), a vision foundation model has significantly advanced in prompt-driven video object segmentation, yet their practical deployment remains limited by the high computational and memory cost of processing dense visual tokens across time. The SAM2 pipelines typically propagate all visual tokens produced by the image encoder through downstream temporal reasoning modules, regardless of their relevance to the target object, resulting in reduced scalability due to quadratic memory–attention overhead. In this work, we introduce a \textbf{text-guided token pruning} framework that improves inference efficiency by selectively reducing token density prior to temporal propagation, without modifying the underlying segmentation architecture. Operating after visual encoding and before memory-based propagation, our method ranks tokens using a lightweight routing mechanism that integrates local visual context, semantic relevance derived from object-centric textual descriptions (either user-provided or automatically generated), and uncertainty cues that help preserve ambiguous or boundary-critical regions. By retaining only the most informative tokens for downstream processing, the proposed approach reduces redundant computation while maintaining segmentation fidelity. Extensive experiments across multiple challenging video segmentation benchmarks demonstrate that post-encoder token pruning provides a practical and effective pathway to efficient, prompt-aware video segmentation, achieving up to \textbf{42.50\%} faster inference and \textbf{37.41\%} lower GPU memory usage compared to the unpruned baseline SAM2, while preserving competitive $\mathcal{J}\&\mathcal{F}$ performance. These results highlight the potential of early token selection to improve the scalability of transformer-based video segmentation systems for real-time and resource-constrained applications.
\end{abstract}

% The projection matrices used for aligning text and uncertainty features with SAM~2 embeddings are computed via closed-form least-squares fitting, preserving the training-free nature of the pipeline.

% \begin{figure}[t]
% \centering
% \includegraphics[width=0.49\textwidth]{tokens_1.png}
% \caption{Qualitative visualisation of retained tokens with the text-driven token pruning approach (on UVO dataset) \cite{wang2021unidentifiedvideoobjectsbenchmark} before passing into SAM2 decoder}
% \label{fig:tokens}
% \end{figure}

%% Graphical abstract
%   \centering
%   % tweak the width if needed: 0.8–1.0\textwidth usually looks good
%   \includegraphics[width=\textwidth]{graphical_abstract.png}

% \begin{highlights}
% \item Introduces a text-guided token pruning module that accelerates SAM-2 without modifying its architecture.
% \item Fuses semantic prompts, uncertainty estimation, and visual context to select only task-relevant tokens before memory propagation.
% \item Achieves up to 42.50\% faster inference and 37.41\% lower GPU memory usage, over baseline SAM2, while preserving segmentation accuracy across five VOS benchmarks.
% \item Demonstrates robustness to prompt noise, reduced human-interaction needs, and broad applicability to real-time and resource-limited settings.
% \end{highlights}

%% Keywords
\begin{keyword}
Interactive video object segmentation \sep Token pruning \sep Segment Anything Model 2 \sep Vision transformers \sep Text-guided segmentation


\end{keyword}

\end{frontmatter}



% \section{Introduction}
% \label{sec1}
% %% Labels are used to cross-reference an item using \ref command.

% Over the past few years we have witnessed a transformative shift in video object segmentation (VOS)~\cite{caelles2017one}, driven by foundation models that bring large-scale pretraining, flexible generalization, and prompt-based control to downstream tasks. Among these, \textbf{Segment Anything Model 2 (SAM~2)}~\cite{ravi2024sam} has emerged as a state-of-the-art engine for interactive video object segmentation (iVOS). SAM~2 supports user inputs such as clicks, boxes, or masks and propagates object masks across long sequences with high temporal consistency. It inherits the architecture of powerful Vision Transformers (ViTs)~\cite{dosovitskiy2020image,ryali2023hiera} from SAM~\cite{kirillov2023segment} and a prompt-guided mask decoder. In contrast to SAM, which focuses on single images, SAM~2 introduces a memory mechanism that stores and reuses past frame information, delivering prompt-sensitive segmentation suitable for robotics, medical workflows, and real-time video editing.

\section{Introduction}

Over the past few years, video object segmentation (VOS) has undergone a transformative shift~\cite{caelles2017one, ravi2024sam}, driven by foundation models that combine large-scale pretraining, flexible generalization, and prompt-based control. Among these, \textbf{Segment Anything Model~2 (SAM~2)}~\cite{ravi2024sam} has emerged as a strong engine for interactive video object segmentation (iVOS). SAM~2 supports sparse user inputs such as clicks, boxes, or masks and propagates object masks across long video sequences with high temporal consistency. Architecturally, it builds upon Vision Transformers (ViTs)~\cite{dosovitskiy2020image,ryali2023hiera} and a prompt-guided mask decoder inherited from SAM~\cite{kirillov2023segment}. Unlike its image-only predecessor, SAM~2 introduces a memory mechanism that stores and reuses visual information across frames, enabling prompt-sensitive segmentation suitable for robotics, medical workflows, and real-time video editing.

\begin{figure}[t]
  \centering
  \includegraphics[width=\linewidth]{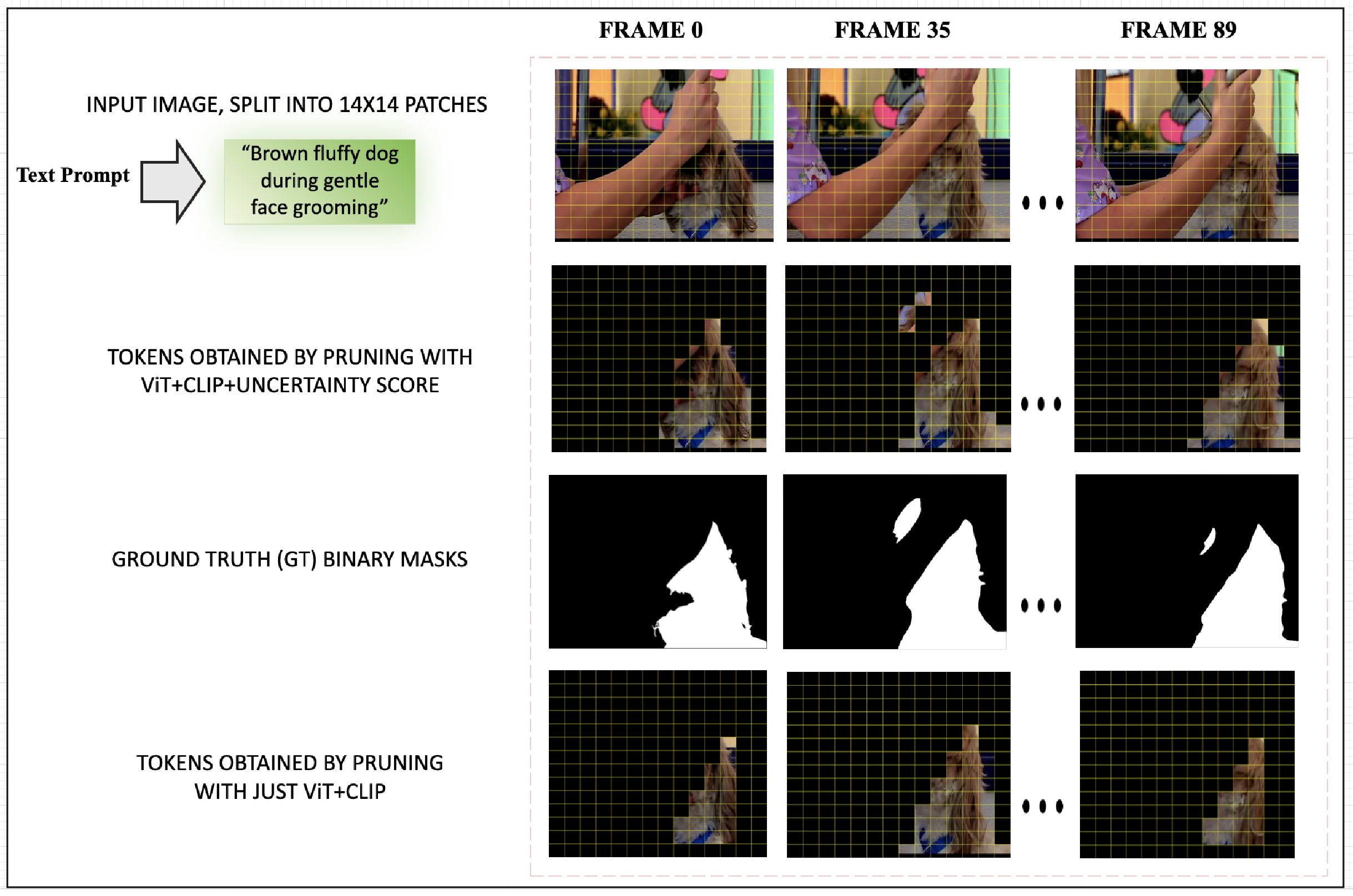}
  \caption{Qualitative visualisation of retained tokens with the text-driven token pruning approach (on UVO dataset) \cite{wang2021unidentifiedvideoobjectsbenchmark} before passing into SAM2 decoder}
  \label{fig1}
\end{figure}

However, SAM~2 and other memory-based VOS systems~\cite{cheng2022xmem,cheng2024putting,zhou2025edgetam,oh2019video,yang2021associating,yang2022decoupling} suffer from a fundamental inefficiency: they treat all visual tokens with equal importance. ViT-based encoders tokenize each frame into dense grids of local patches, which are processed by multi-headed self-attention \cite{vaswani2017attention} and deep transformer stacks. In SAM~2, these tokens are further stored and repeatedly accessed by the memory engine across time. Regardless of whether a token corresponds to static background, ambiguous boundaries, or semantically relevant foreground, it is carried through the same attention-heavy propagation pipeline. As a result, memory usage and inference latency increase substantially with the number of stored tokens, creating a bottleneck for latency-sensitive and resource-constrained settings.

\begin{figure}[t]
  \centering
  \includegraphics[width=\linewidth]{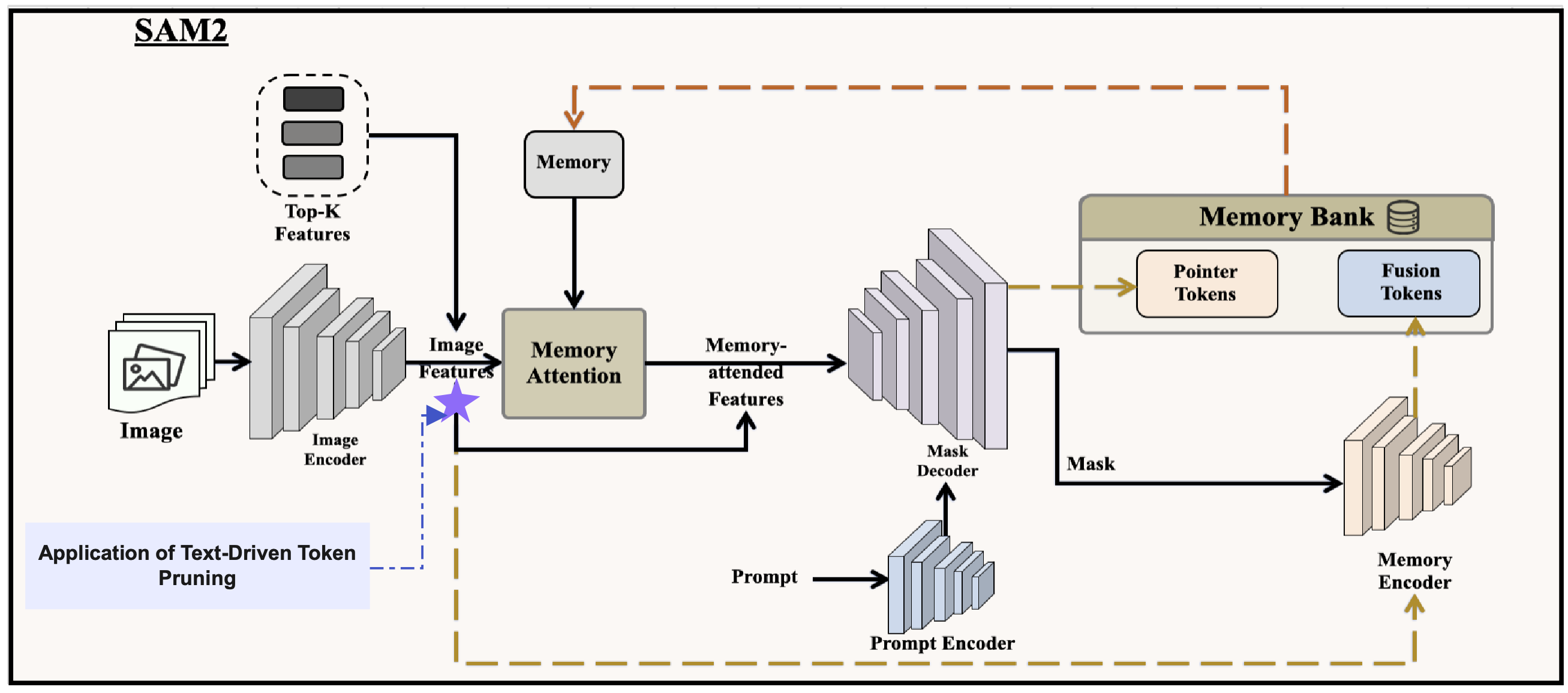}
  \caption{Qualitative visualization steps of segmentation with our text-driven token pruning approach atop SAM2}
  \label{fig2}
\end{figure}

A straightforward way to address this issue would be to sparsify \cite{luo2018convolutional} the image encoder itself. However, the SAM~2 encoder is typically used in a frozen, pretrained form, and modifying its internal token flow would require retraining or architectural changes. In this work, we therefore take a different approach: rather than altering the encoder, we propose a \textbf{post-encoder, text-driven token pruning framework} (as illustrated in Figure.~\ref{fig2}) that operates \emph{between} the image encoder and the memory/mask-decoder stack. Our method selectively filters the encoder’s output tokens before they enter SAM~2’s memory and temporal attention modules, yielding substantial computational and memory savings where they matter most, while remaining non-invasive to the backbone.

To decide which tokens should be retained, we incorporate two additional sources of information alongside the visual embeddings: \textbf{semantic cues derived from text} \cite{radford2021learning} and \textbf{model uncertainty over ambiguous regions} \cite{kendall2018uncertainty, kendall2015bayesian, liu2024uncertainty}. Intuitively, as illustrated in Figure.~\ref{fig1} only a subset of tokens is relevant to a given segmentation intent (e.g., ``segment the brown fluffy dog during gentle face grooming''), while other regions may require preservation due to visual ambiguity even if they are weakly aligned semantically. We therefore evaluate tokens using three complementary signals—semantic relevance, predictive uncertainty, and visual context—which are fused to retain only the most informative subset per frame. This strategy reduces redundant memory accumulation, improves inference speed, and preserves segmentation accuracy over long temporal horizons.

Semantic relevance is obtained by aligning each visual token with a compact text embedding that reflects the intended object or region of interest. This text may be provided directly by a user, or automatically generated when no explicit instruction is available. In the latter case, we employ a lightweight two-stage procedure in which a vision–language model proposes an object-centric description from a coarse region of interest, followed by a refinement step that distills it into a concise phrase. We emphasize that automatically generated text does not guarantee perfect alignment with user intent; rather, it serves as a semantic prior. Accordingly, we study robustness to vague, partial, or overly verbose prompts and show that the pruning mechanism remains stable under such noise (shown in table~\ref{tab:prompt_ablation}).

Predictive uncertainty is estimated via Monte Carlo Dropout~\cite{kendall2015bayesian, kendall2018uncertainty, liu2024uncertainty} applied to intermediate transformer layers. Tokens that exhibit high variability across stochastic forward passes typically correspond to edges, occlusions, motion blur, or visually confusing regions—precisely those that should not be pruned aggressively. We empirically select intermediate layers that balance local detail and contextual information \cite{raghu2021vit, liu2021swin, kendall2018uncertainty}, and we include ablations that isolate the contribution of uncertainty (table \ref{tab:ablation_signals}) and vary the number of Monte Carlo passes to quantify both accuracy gains and efficiency trade-offs (table \ref{tab:mcd}).

The resulting token scores are fused using a lightweight MLP \cite{rumelhart1986learning, almeida2020multilayer} that ranks tokens by importance. Only the top-$k$ tokens per frame are retained and passed to the SAM~2 memory and mask decoder. Importantly, SAM~2 already operates on a variable-length set of memory tokens, so reducing this sequence length requires no architectural modification. For initialization, we follow the standard iVOS protocol and use a single positive click (point) in the first frame containing the target object. In practical scenarios, this click may be provided by a user or computed automatically; for objects with complex shapes or holes, we adopt a distance-transform-based representative click (point) rather than a naive geometric centroid, which better reflects typical human interactions.

Compared to prior token pruning methods~\cite{tang2023dynamic,liu2024revisiting}, our approach is the first to integrate semantic alignment and uncertainty modeling into token selection for a memory-based iVOS pipeline. Earlier approaches typically rely on heuristic saliency measures, operate on single images, or require retraining the backbone. In contrast, our method is \emph{training-free at segmentation time}: the projection operators used for semantic and uncertainty alignment are obtained via closed-form fitting on encoder features. The only gradient-based optimization occurs in the lightweight MLP, which routes the semantically aligned tokens and prunes irrelevant ones, after the image encoder and before entering the memory stack for segmentation. This makes the proposed framework lightweight, modular, and easy to integrate with existing SAM~2-based systems.

We evaluate our approach on five challenging benchmarks—UVO~\cite{wang2021unidentifiedvideoobjectsbenchmark}, PUMAVOS~\cite{bekuzarov2023xmempp, Bekuzarov_2023_ICCV}, EndoVis~\cite{allan20202018, allan2021stereo}, VOST~\cite{tokmakov2023breaking}, and LVOSv2~\cite{hong2025lvos, hong2024lvos}—and compare against strong VOS baselines including STM, AOT, DeAOT, XMem, Cutie, and SAM~2. Across datasets, we observe up to \textbf{$37.41\%$ reduction in GPU memory usage} and \textbf{$42.50\%$ faster inference} relative to SAM~2, while preserving competitive $\mathcal{J}$\&$\mathcal{F}$ accuracy. Additional analyses examine the trade-off between token retention rate and performance, sensitivity to uncertainty estimation, and robustness to imperfect prompts and out-of-distribution imagery.

\noindent\textbf{Our key contributions are summarized as follows:}
\begin{itemize}
    \item Our primary goal is not to introduce a new segmentation architecture, but to improve the deployability, efficiency, and robustness of foundation segmentation models in \textbf{real-world video settings}.
     \item We propose a text-guided token pruning framework for interactive video object segmentation that integrates \textbf{semantic alignment, uncertainty estimation, and visual context} to rank and select task-relevant tokens prior to temporal propagation.
    \item Our approach is a modular, post image encoder pruning design that can be seamlessly integrated into SAM~2 \textbf{reducing memory and attention overhead} in the downstream propagation and decoding stages.
\end{itemize}

\section{Related Works}
\label{sec2}

In this section, we review prior work in interactive video object segmentation, foundation models for image and video segmentation, and token-level pruning for efficient Vision Transformers. We aim to situate our method within these lines of research, clarifying both architectural context and practical constraints that motivate token pruning between the encoder and propagation stages of SAM~2.

\subsection{Interactive Video Object Segmentation (iVOS)}
Interactive video object segmentation (iVOS) refers to identifying and propagating object masks across video frames with sparse human input such as clicks, boxes, or partial masks. Classical approaches such as STM~\cite{oh2019video}, AOT~\cite{yang2021associating}, and XMem~\cite{cheng2022xmem} use spatio-temporal memory mechanisms to match query frames against stored object embeddings. These methods achieve strong performance but often require repeated interaction or complex memory updates.

Foundation models have reshaped this landscape. SAM~\cite{kirillov2023segment} introduced prompt-driven segmentation for images, while SAM~2~\cite{ravi2024sam} extended this paradigm to videos through a long-range memory engine. SAM~2 supports both interactive refinement and semi-automatic propagation, but it inherits a computational bottleneck: its memory stack stores and attends to \emph{dense} token grids from every processed frame, regardless of their informativeness. Recent on-device variants such as EdgeTAM~\cite{zhou2025edgetam} and lightweight propagation frameworks focus on architectural compression or approximated attention mechanisms, yet they preserve dense token flows.

% Grounding-based extensions such as Grounded-SAM or Grounded-SAM~2 integrate text–region alignment via pretrained grounding modules. However, these approaches introduce additional heads, require training on grounding annotations, and do not alter the dense token propagation in SAM~2's memory. Because our goal is to accelerate inference without architectural modifications or retraining, we do not directly adopt these grounding variants. This motivates a design that uses text embeddings only as \emph{guidance signals} for ranking encoder tokens rather than as mask-generating modules.

\subsection{Foundation Models for Efficient VOS}
A parallel research thread aims to make VOS systems more computationally efficient. Variants such as DeAOT~\cite{yang2022decoupling}, Cutie~\cite{cheng2024putting}, and efficient long-term trackers reduce propagation overhead using compact memories or hierarchical matching. While these models reduce computation in the propagation module, they still rely on dense spatial token grids at the feature-encoding stage. None explicitly address the redundancy in token sets \emph{before} memory insertion, which becomes significant in long videos where memory grows linearly with time.

Our method is complementary to these approaches: instead of redesigning SAM~2’s \cite{ravi2024sam} architecture, we intervene at the token level by filtering encoder outputs before they reach the memory bank. Thus, our pruning strategy can be combined with many of these efficient propagation methods.

\subsection{Token Pruning in Vision Transformers}
Vision Transformers tokenize inputs into spatial patches and apply quadratic-cost attention over all tokens~\cite{dosovitskiy2020image}. As a result, pruning or compressing tokens has become a popular strategy for accelerating both classification and dense prediction models. Early methods such as DynamicViT~\cite{rao2021dynamicvitefficientvisiontransformers} and TokenLearner~\cite{ryoo2021tokenlearner} learn to identify salient tokens, while later approaches focus on task-specific regimes such as detection or instance segmentation~\cite{liu2024revisiting,liu2023revisitingtokenpruningobject}. These works demonstrate that informative tokens often occupy a small subset of the spatial grid.

However, few gaps remain.  
First, most pruning methods require \emph{retraining the backbone}, which is incompatible with SAM~2 where frozen encoder weights are typically used.  
Second, existing pruning strategies operate primarily on visual saliency or attention magnitude; they do not incorporate semantic intent from prompts or predictive uncertainty to preserve ambiguous regions.

Recent multimodal pruning methods~\cite{chen2025vltp,shrivastava2023clip} investigate text-guided filtering for image segmentation, but they do not address video memory, do not integrate uncertainty, and are not directly compatible with SAM~2’s inference interfaces.

iVOS foundations (SAM~2 and its predecessors) provide strong segmentation quality but propagate dense tokens through costly memory operations; efficient-VOS methods reduce propagation cost but do not address token redundancy; and token pruning literature introduces token sparsity but assumes retraining or lacks semantic and uncertainty conditioning.  
Our work bridges these lines by proposing a \emph{post-encoder, text-guided and uncertainty-aware} token pruning module compatible with SAM~2’s architecture, intended to reduce the computational footprint of its memory and decoding stack without modifying the encoder or requiring fine-tuning.

\begin{figure}[t]
  \centering
  \includegraphics[width=\linewidth]{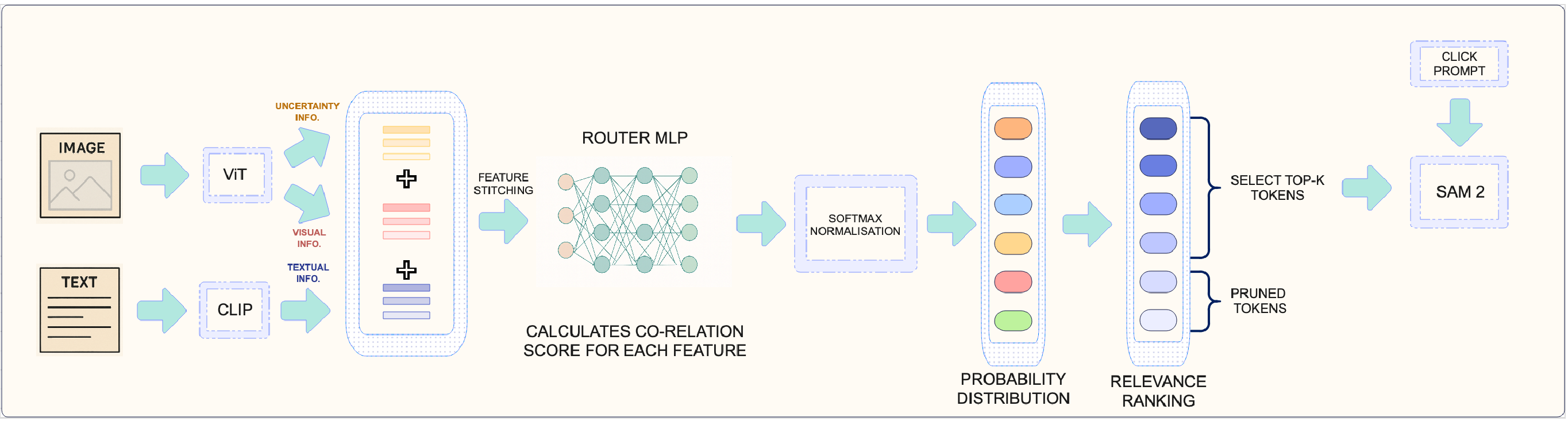}
  \caption{End-to-end pipeline of our text-driven token pruning framework atop SAM2. Semantic cues from text prompts, visual context from ViT tokens,
along with token uncertainty are fused by a lightweight router to retain only task-relevant tokens before SAM2 decoding.}
  \label{fig3}
\end{figure}

\begin{figure}[t]
  \centering
      \includegraphics[width=\linewidth]{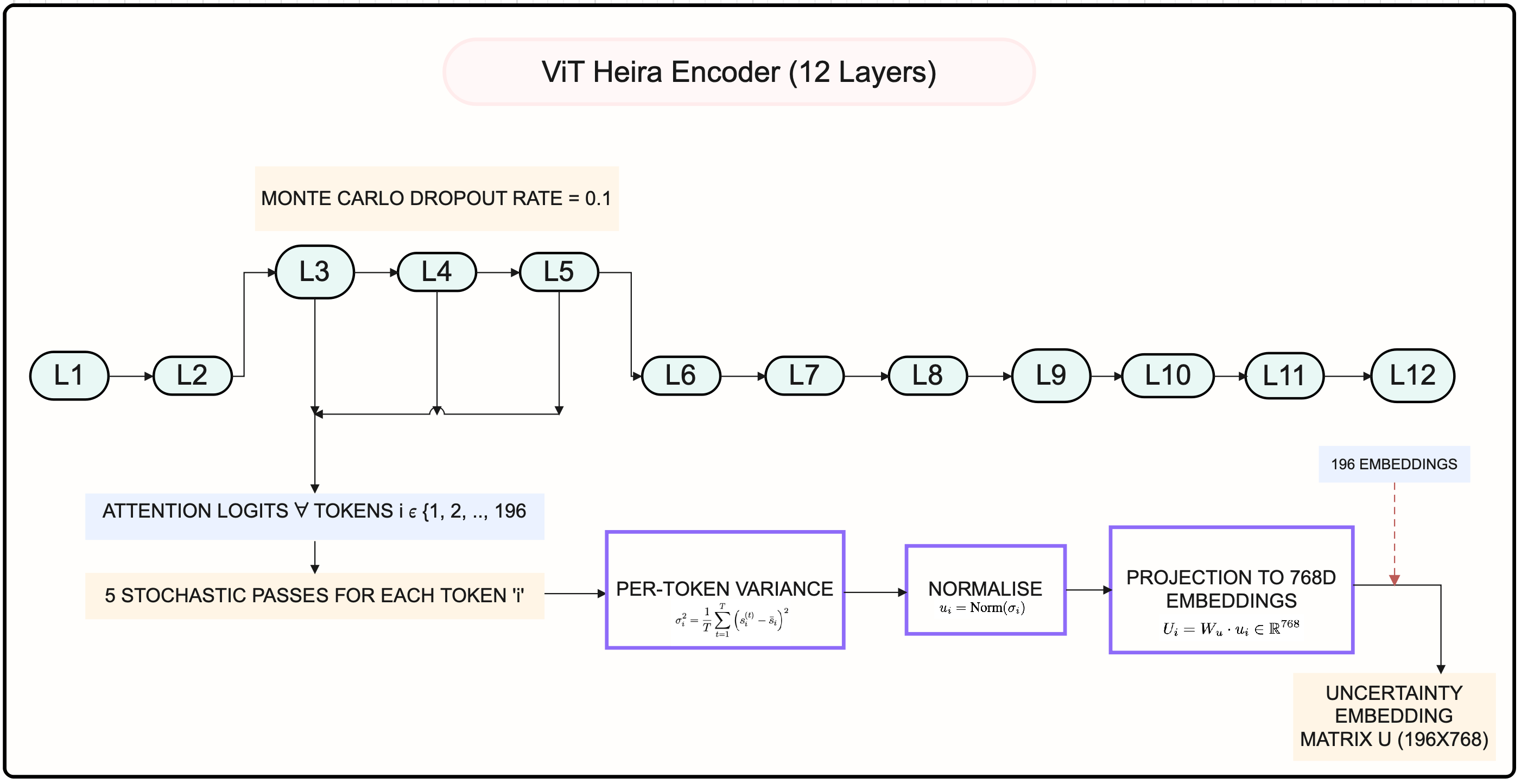}
  \caption{Monte Carlo (MC) Dropout applied to ViT-Hiera layers L3–L5 for uncertainty prediction per token}
  \label{fig4}
\end{figure}

\section{Method}
\label{sec3}

\subsection{Overview}
Our goal is to accelerate SAM~2 by reducing the number of visual tokens processed in its \emph{memory} and \emph{mask-decoder} pathways, which constitute the dominant inference-time bottlenecks. Importantly, our method does \emph{not} prune tokens inside the ViT-Hiera encoder—which remains dense and frozen for accurate multi-headed self attention—but instead intervenes \emph{after} the encoder and \emph{before} tokens are inserted into the memory engine. As illustrated in Fig.~\ref{fig3}, the proposed module computes semantic relevance, predictive uncertainty, and visual-context features for all tokens, ranks them using a lightweight scoring network, and selects the top-$k$ tokens per frame:
\begin{equation}
X_{\text{pruned}} = \text{TopK}\!\big(f_{\theta}(X_{\text{ViT}},\, e_{\text{text}},\, U)\big).
\label{eq:topk}
\end{equation}
The pruned tokens are used throughout SAM~2’s propagation and decoding stack without requiring architectural modification, since the memory engine already supports variable-length token sequences.

\begin{figure}[t]
  \centering
  \includegraphics[width=\linewidth]{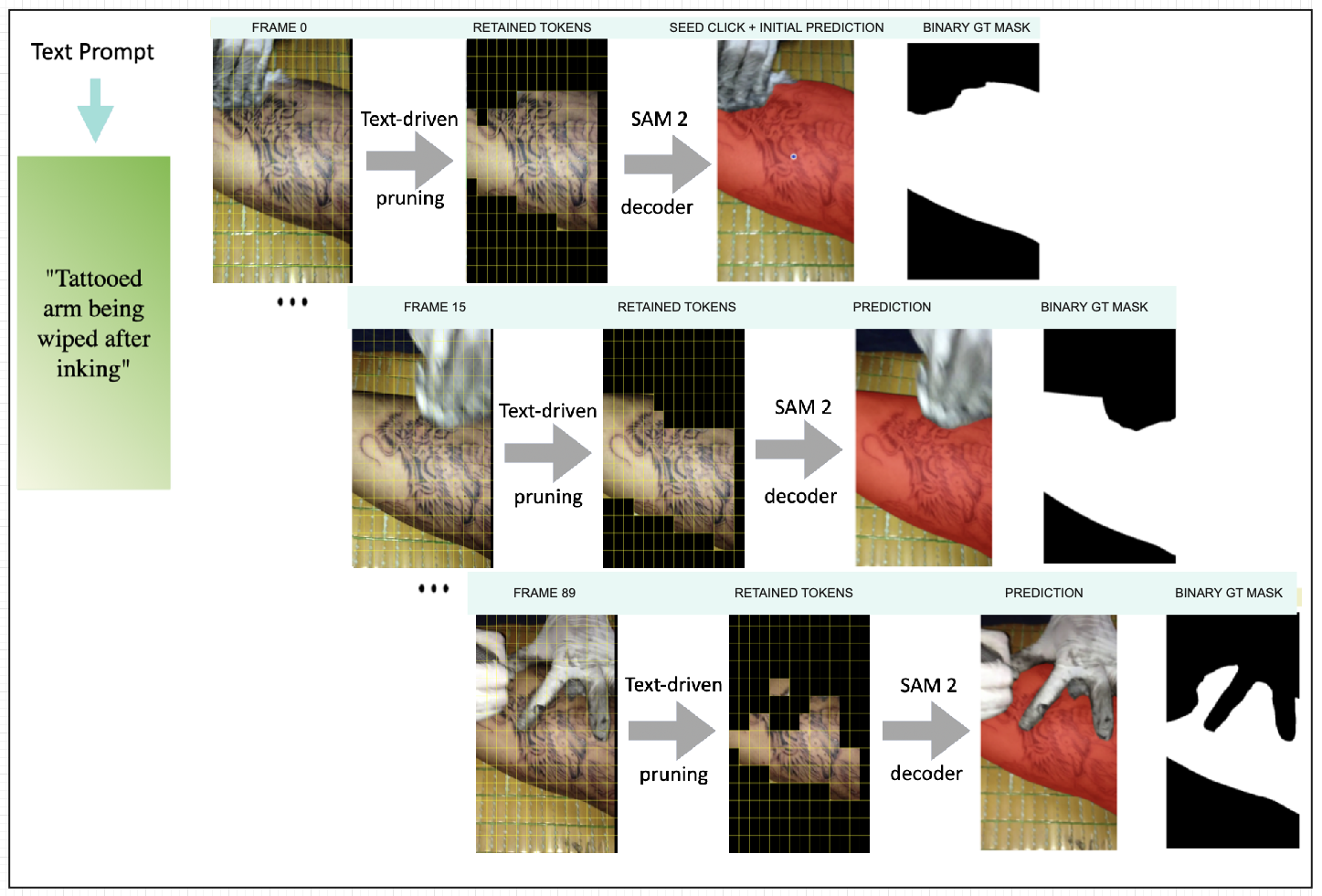}
  \caption{Qualitative segmentation results on a UVO dataset~\cite{wang2021unidentifiedvideoobjectsbenchmark} sequence with our text-driven token pruning approach atop SAM2. We achieve mean $\mathcal{J\&F}$ of 97.95\% with no other refinement clicks required other than seed click to serve as input to SAM2's prompt encoder. The image shows the dense tokens in the input frame, followed by semantically relevant tokens at 30\% retention, and the segmentation mask by SAM2, with the ground truth also shown for comparison.}
  \label{fig5}
\end{figure}

\vspace{2mm}
\subsection{Visual Token Extraction from SAM~2 Encoder}
Each frame $I_t$ is processed by SAM~2’s frozen ViT-Hiera encoder, producing a dense token grid:
\begin{equation}
X_{\text{ViT}} = \text{ViT-Hiera}(I_t) \in \mathbb{R}^{N \times d_v}, \quad N = 14 \times 14 = 196.
\label{eq:vit}
\end{equation} Here, $X_{\text{ViT}} \in \mathbb{R}^{N \times d_v}$ denotes the visual token embeddings produced by the SAM~2 image encoder, with $N=196$ and $d_v=768$, the typical dimension of ViT visual tokens.
These token embeddings are the only feature maps available to all downstream SAM~2 modules; we therefore prune strictly \emph{after} this stage.

\vspace{2mm}
\subsection{Semantic Prompt Extraction Without Ground Truth}
\label{sec:prompt}

In realistic deployment, ground-truth masks are unavailable to specify segmentation intent. Thus, we eliminate ground-truth dependency in the prompt-generation step.  
A user may optionally provide a short textual instruction.  
When no user-provided text is available, we obtain a coarse region of interest using a class-agnostic object proposal mechanism. Specifically, we deploy a Region Proposal Network (RPN), Faster R-CNN \cite{ren2015faster}, to generate a set of candidate bounding boxes, each associated with an objectness score. We select the proposal with the highest objectness score as the foreground region of interest in the first frame. A vision–language model such as, Large Language and Vision Assistant (LLaVA)~\cite{liu2023visual, lin2023video, lin2024video} generates a coarse caption for the region, as in case of fig~\ref{fig1} we obtained description as, "A small fluffy dog with brown fur
is being gently held by a person’s
hand holding grooming tools around
it’s face during grooming". This caption is then refined by a small language model, Bidirectional Encoder Representations from Transformers (BERT) \cite{devlin2019bert, koroteev2021bert} into a concise phrase, "Brown fluffy dog during gentle face
grooming" (prompt used in fig~\ref{fig1}). This provides a lightweight semantic prior, with minimal +0.5s overhead, when no user intent is available in a realistic deployment (practically feasible with any underlying segmentation pipeline).

\vspace{1mm}

% ---------------- Original text (UNCHANGED) ----------------
We encode the prompt using a frozen CLIP \cite{radford2021learning} text encoder:
\begin{equation}
e_{\text{text}} = f_{\text{CLIP}}(\mathcal{P}) \in \mathbb{R}^{d_t}.
\label{eq:clip}
\end{equation} where, $\mathcal{P}$ denotes the input text prompt; $f_{\text{CLIP}}(\cdot)$ is a frozen CLIP text encoder; $e_{\text{text}} \in \mathbb{R}^{d_t}$ is the resulting text embedding with $d_t=512$.
To match SAM~2’s visual token dimensionality ($d_v = 768$), we compute a \emph{training-free} least-squares projection:
\begin{equation}
W_t = \arg\min_{W \in \mathbb{R}^{d_t \times d_v}} \| X_{\text{ViT}} W - e_{\text{text}} \|_2^2,
\label{eq:Wt}
\end{equation} where, $W_t$ is training-free least-squares projection matrix;
 $e_{\text{text}}$ is the text vector broadcast across tokens. This projection is computed once per video, requires no labels, and preserves the training-free nature of the method. The aligned semantic embedding is:
\begin{equation}
e'_{\text{text}} = W_t^\top e_{\text{text}} \in \mathbb{R}^{768}.
\label{eq:etextprime}
\end{equation}

\vspace{2mm}
\subsection{Token-Level Uncertainty Estimation}
\label{sec:uncertainty}

Ambiguous regions such as occlusions, motion blur, or thin structures are precisely estimated  where aggressive pruning is harmful. To preserve such regions, we estimate predictive uncertainty via Monte Carlo Dropout (MCD) \cite{kendall2018uncertainty} applied to intermediate encoder layers, as shown in Figure.~\ref{fig4}. Following empirical studies on ViTs, we apply dropout to layers 3--5 \cite{raghu2021vit, liu2021swin, kendall2018uncertainty}, which offer strong spatial detail while maintaining contextual awareness.

For each of $T$ stochastic forward passes, we extract the pre-softmax attention logits $s_i^{(t)}$ for token $i$:
\begin{equation}
s_i^{(t)} = f_{\text{attn}}^{(t)}(X_{\text{ViT}}), \quad t = 1,\dots,T.
\label{eq:attn}
\end{equation}
Uncertainty is quantified as variance:
\begin{equation}
\sigma_i^2 = \frac{1}{T} \sum_{t=1}^{T} \left(s_i^{(t)} - \bar{s}_i\right)^2, 
\qquad 
\bar{s}_i = \frac{1}{T} \sum_{t=1}^{T} s_i^{(t)}.
\label{eq:unc}
\end{equation} where, $\sigma_i^2$ denotes the variance-based uncertainty estimate for token $i$, $\bar{s}_i$ denotes the mean attention logit for token \emph{i} over all \emph{T} stochastic passes.

% ----------- Added normalization equation (ONLY ADDITION) -----------
The standard deviation $\sigma_i$ values are normalized across tokens using min--max normalization:
\begin{equation}
\tilde{\sigma}_i = \frac{\sigma_i - \min_j \sigma_j}{\max_j \sigma_j - \min_j \sigma_j}.
\label{eq:unc_norm}
\end{equation} where, $\tilde{\sigma}_i$ is its normalized form

After normalization, uncertainty values are projected to token space using a second least-squares projection:
\begin{equation}
W_u = \arg\min_{W} \left\| X_{\text{ViT}} W - \tilde{\sigma}_i \right\|_2^2, 
\qquad 
U_i = W_u^\top \tilde{\sigma}_i.
\label{eq:Wu}
\end{equation} where, $W_u$ is training-free least-squares projection matrix; $U_i$ is the uncertainty feature aligned to the visual token space. 
This projection also requires no training signals.  
We further benchmark $T \in \{4,5,6\}$ to quantify the efficiency–accuracy trade-off.

\vspace{2mm}
\subsection{Fused Token Representation}
\label{sec:fusion}

For token $i$, we fuse visual features, aligned semantic embedding, and uncertainty features:
\begin{equation}
h_i = [\, X_{\text{ViT}, i} \, ; \, e'_{\text{text}} \, ; \, U_i \, ] 
\in \mathbb{R}^{3d_v}.
\label{eq:hi}
\end{equation} where, $h_i$ denotes the fused token representation

\vspace{2mm}
\subsection{Token Scoring and Pruning}
Each fused descriptor is passed through a lightweight two-layer MLP:
\begin{equation}
s_i = \text{MLP}(h_i), \qquad 
\text{MLP}: 2304 \rightarrow 256 \rightarrow 1.
\label{eq:mlp}
\end{equation} 
[Note that the input size to the MLP is 2304 vectors, as we have 3 signals (visual, semantic, and uncertainty) each aligned to 768 dimensions now.]
Scores are softmax-normalized:
\begin{equation}
\alpha_i = \frac{\exp(s_i)}{\sum_{j=1}^{N} \exp(s_j)}.
\label{eq:softmax}
\end{equation} where, $s_i$ and $\alpha_i$ are the raw and normalized token importance scores, respectively.
We retain the top-$k$ tokens:
\begin{equation}
X_{\text{pruned}} = \{ X_{\text{ViT}, i} \,|\, i \in \text{TopK}(\alpha, k) \}.
\label{eq:pruned}
\end{equation}

SAM~2’s memory accepts variable-length sequences, so no architectural modification is required. The pruned tokens simply replace the full set of 196 tokens for memory writing and mask decoding.

\FloatBarrier
\begin{figure}[t]
  \centering
  \includegraphics[width=\linewidth]{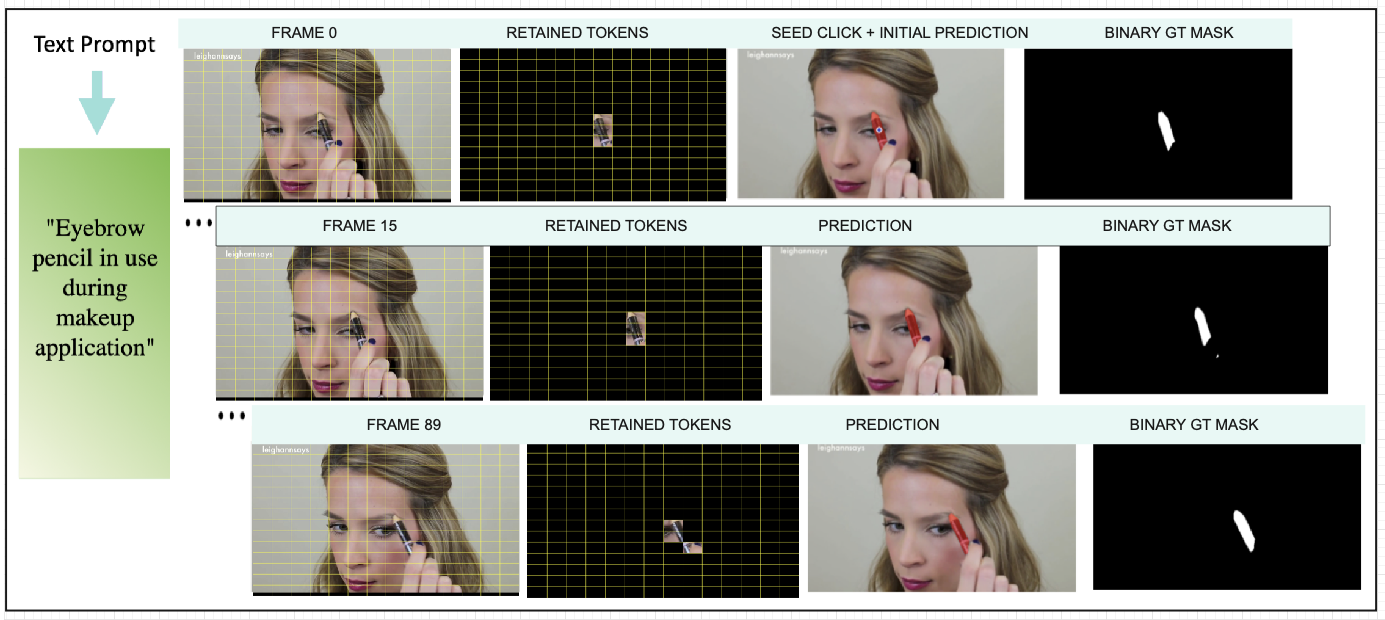}
  \caption{Qualitative segmentation results on a UVO dataset~\cite{wang2021unidentifiedvideoobjectsbenchmark} sequence with our text-driven token pruning approach atop SAM2. On this video sequence, we achieve mean $\mathcal{J\&F}$ of 91.84\% and minimum $\mathcal{J\&F}$ of 73.85\% with just 4 refinement clicks required including the seed click to serve as input to SAM2's prompt encoder. The image shows the dense tokens in the input frame, followed by semantically relevant tokens at 30\% retention, and the segmentation mask by SAM2, with the ground truth binary mask also shown for qualitative comparison.}
  \label{fig6}
\end{figure}

\begin{figure}[t]
  \centering
  \includegraphics[width=\linewidth]{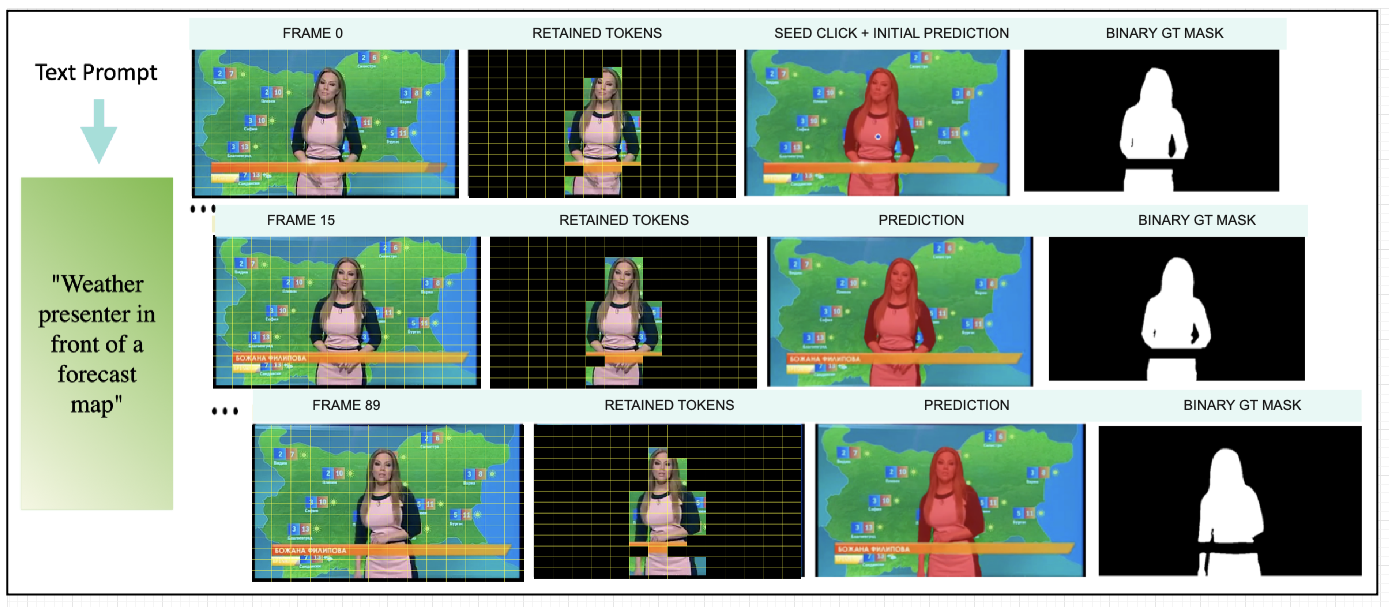}
  \caption{Qualitative segmentation results on a UVO dataset~\cite{wang2021unidentifiedvideoobjectsbenchmark} sequence with our text-driven token pruning approach atop SAM2. On this video sequence, we achieve mean $\mathcal{J\&F}$ of 96.95\% and minimum $\mathcal{J\&F}$ of 96.01\% with no other refinement clicks required other than the seed click to serve as input to SAM2's prompt encoder. The image shows the dense tokens in the input frame, followed by semantically relevant tokens at 30\% retention, and the segmentation mask by SAM2, with the ground truth binary mask also shown for qualitative comparison.}
  \label{fig7}
\end{figure}
\FloatBarrier

\vspace{2mm}
\subsection{Integration With SAM~2 Memory and Prompt Encoder}
\label{sec:integration}

A single positive click is required as input for the prompt encoder to initialize SAM~2.  
We propose an initialization click prompt algorithm either as (i) a user click or (ii) a distance-transform–based representative point for irregular shapes (e.g., torus-like objects). This avoids incorrect geometric centroid selection.

Let $\mathbf{p}$ denote the initial click, which is embedded via SAM~2’s prompt encoder:
\begin{equation}
z_{\text{prompt}} = f_{\text{prompt}}(\mathbf{p}).
\label{eq:prompt}
\end{equation}

At each frame, the mask is produced by SAM~2’s decoder:
\begin{equation}
M_t = f_{\text{decoder}}(X_{\text{pruned}},\, z_{\text{prompt}},\, \mathcal{M}_{1:t-1}),
\label{eq:mask}
\end{equation}
where $\mathcal{M}_{1:t-1}$ denotes the memory bank.

Because the memory engine accumulates only pruned tokens rather than full dense grids, its computational and memory cost reduce proportionally.

\vspace{2mm}
\subsection{Optional Interactive Refinement}
Although our method does not require high human-in-the-loop support, SAM~2 inherently allows optional manual refinement. If $\mathcal{J\&F}$ drops below a threshold, a synthetic or user-provided click can be added, leading to semi-automatic propagation, core in interactive video object segmentation. We observe empirically that pruning reduces drift and therefore reduces refinement calls.

\begin{algorithm}[t]
\caption{Fast SAM2 with Text-Driven Token Pruning}
\label{alg:token_pruning}
\small
\KwIn{Video frames $\{I_t\}_{t=1}^{L}$; optional text prompt $\mathcal{P}$; token budget $k$; MC passes $T=5$}
\KwOut{Segmentation masks $\{M_t\}_{t=1}^{L}$}

Encode text prompt using frozen CLIP:\;
$e_{\text{text}} \leftarrow f_{\text{CLIP}}(\mathcal{P})$\;

Compute semantic alignment:\;
$W_t \leftarrow \arg\min_{W}\|X_{\text{ViT}}W - e_{\text{text}}\|_2^2$\;
$e'_{\text{text}} \leftarrow W_t^\top e_{\text{text}}$\;

\For{$t = 1$ \KwTo $L$}{
    $X_{\text{ViT}} \leftarrow \text{ViT-Hiera}(I_t)$\;

    \For{$\tau = 1$ \KwTo $T$}{
        $s_i^{(\tau)} \leftarrow f_{\text{attn}}^{(\tau)}(X_{\text{ViT}})$\;
    }

    $\sigma_i^2 \leftarrow \frac{1}{T}\sum_{\tau}(s_i^{(\tau)}-\bar{s}_i)^2$\;
    $\tilde{\sigma}_i \leftarrow \frac{\sigma_i-\min_j\sigma_j}{\max_j\sigma_j-\min_j\sigma_j}$\;

    $W_u \leftarrow \arg\min_{W}\|X_{\text{ViT}}W-\tilde{\sigma}\|_2^2$\;
    $U_i \leftarrow W_u^\top \tilde{\sigma}_i$\;

    $h_i \leftarrow [X_{\text{ViT},i};\ e'_{\text{text}};\ U_i]$\;
    $s_i \leftarrow \text{MLP}(h_i)$\;
    $\alpha_i \leftarrow \frac{\exp(s_i)}{\sum_j\exp(s_j)}$\;

    $X_{\text{pruned}} \leftarrow \text{TopK}(\alpha,k)$\;
    $M_t \leftarrow f_{\text{decoder}}(X_{\text{pruned}}, z_{\text{prompt}}, \mathcal{M}_{1:t-1})$\;
}

\Return $\{M_t\}_{t=1}^{L}$
\end{algorithm}

\section{Experiments}
\label{sec4}

All experiments are conducted using an NVIDIA GeForce RTX 3090 GPU. SAM2 is initialized using official ViT-Hiera (tiny) checkpoint~\cite{ryali2023hiera}. Experiments are conducted using images resized to an input resolution of $224 \times 224$. The visual encoder (ViT-Hiera~\cite{ryali2023hiera}) produces 14x14 spatial tokens, each with a feature dimension of 768. Textual prompts distilled from LLaVA~\cite{liu2023visual} and BERT~\cite{devlin2019bert} are projected into a 512-dimensional embedding space. To estimate uncertainty, we perform $T=5$ stochastic forward passes using Monte Carlo dropout applied in layers 3--5 of SAM2's visual encoder. The router MLP~\cite{tolstikhin2021mlp, rumelhart1986learning, haykin1994neural, almeida2020multilayer} used for token scoring consists of two linear layers separated by a GELU~\cite{hendrycks2016gaussian, zhang2018gelu} activation. Following scoring, we retain the top 30\% of tokens for downstream propagation. For interactive segmentation, we initialize SAM2 with a single positive point prompt (user initialized or automated) located at the centroid of the object of interest in the first frame that contains this object. This serves as the initial condition for the memory encoder. In our experiments, we allow a maximum of 10 refinements rounds per sequence with 90 frames (including starter click), with up to 3 clicks allowed in each refinement. Figures.~\ref{fig5}, \ref{fig6}, \ref{fig7} illustrate the \textbf{qualitative visualisation} of results on several videos by leveraging our text-driven token pruning to accelerate SAM2.

Best and second-best results are highlighted in \textbf{green} and \textbf{yellow} shades respectively, in all of tables \ref{tab:main_results}-\ref{tab:prompt_ablation}.

\subsection{Evaluation Metrics}
We adopt the standard $\mathcal{J\&F}$ metric~\cite{pont20172017, perazzi2016benchmark}, which averages region similarity using the Jaccard index ($\mathcal{J}$) and contour accuracy using the F-measure ($\mathcal{F}$). Unless otherwise specified, scores are averaged over all annotated frames and all object IDs and reported per dataset.

\begin{equation}
\mathcal{J\&F} = \frac{1}{2TO_t} \sum_{t=1}^{T} \sum_{o=1}^{O_t} \left( \frac{|S_{t,o} \cap G_{t,o}|}{|S_{t,o} \cup G_{t,o}|} + \frac{2P^{c}_{t,o} R^{c}_{t,o}}{P^{c}_{t,o} + R^{c}_{t,o}} \right)
\end{equation}

\noindent
where $S$, $G$, $P^{c}$, $R^{c}$, $T$, and $O_t$ refer to the output mask, ground-truth mask, precision between output and ground-truth contours, recall between output and ground-truth contours, number of frames, and number of objects in each frame, respectively.

\subsection{Main Results}
\begin{figure}[h]
  \centering
  \includegraphics[width=\linewidth]{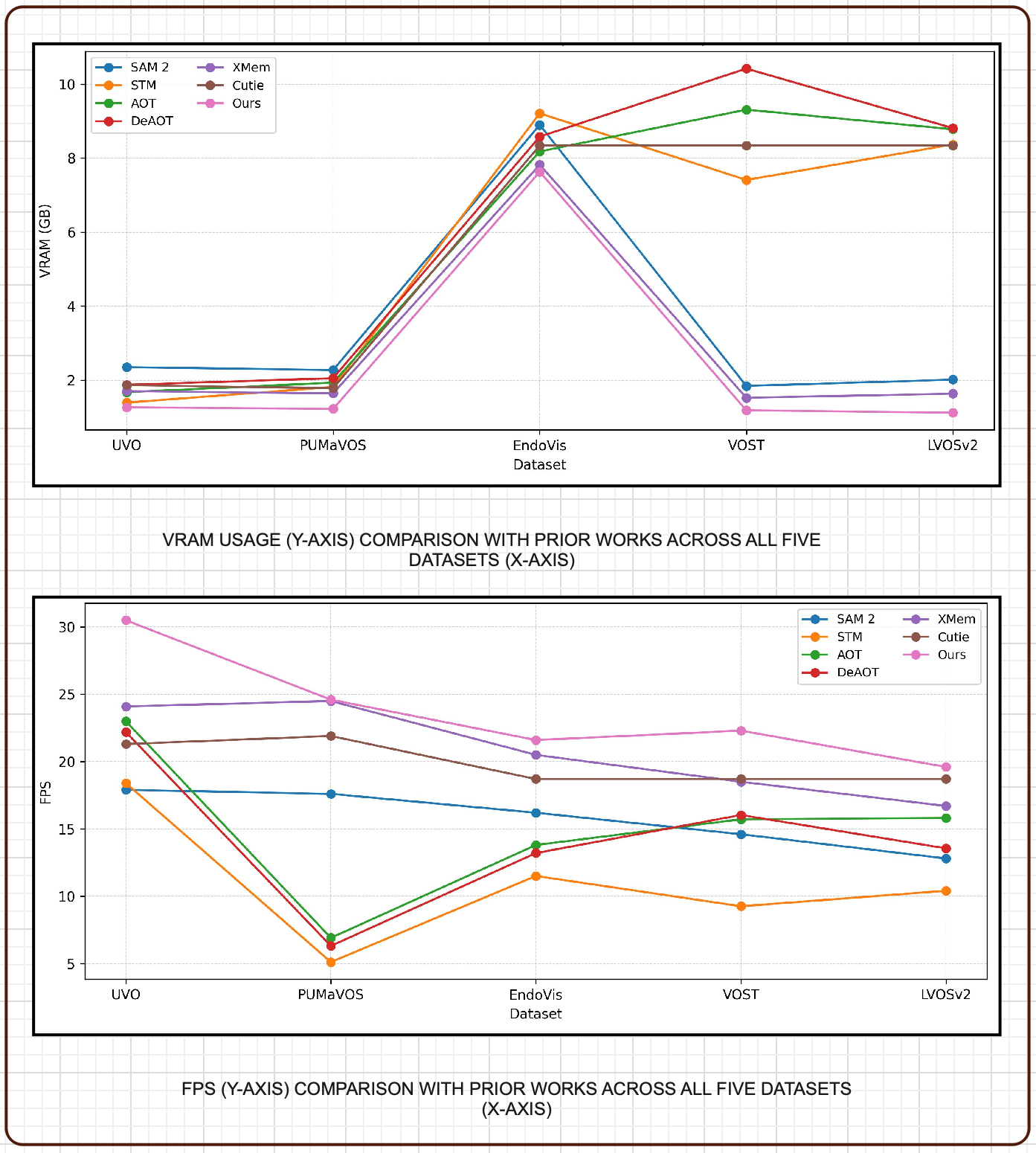}
  \caption{Efficiency comparison across five video object segmentation benchmarks.
Top: GPU memory consumption (VRAM). Bottom: inference speed (FPS).
Our text-guided token pruning consistently reduces memory usage and improves runtime efficiency compared to prior methods, including SAM2, without altering the underlying segmentation architecture. The results demonstrate that selectively propagating only semantically and visually relevant tokens substantially improves scalability across diverse datasets.}
  \label{fig9}
\end{figure}

To evaluate the performance of our method, we conduct experiments on several benchmark datasets with dense annotation (all sets of PUMaVOS~\cite{Bekuzarov_2023_ICCV} and EndoVis2018~\cite{allan2021stereo}, as well as validation sets of VOST~\cite{tokmakov2023breaking}, UVO~\cite{wang2021unidentifiedvideoobjectsbenchmark}, and
LVOSv2~\cite{hong2024lvos}).
We also include the evaluation of the baseline SAM2 and five prior SOTA methods in VOS task, STM~\cite{oh2019video}, AOT~\cite{yang2021associating}, DeAOT~\cite{yang2022decoupling}, XMem~\cite{cheng2022xmem}, and Cutie~\cite{cheng2024putting}.
For simplicity, we only perform segmentation on objects that appear in the first frame.
Table~\ref{tab:main_results} and figure~\ref{fig9} show that in terms of inference speed and memory consumption, our method yield competitive results against prior SOTA models (averaged) by an average frames per second (FPS) increase of \textbf{31.7\%} and average GPU memory usage reduction of \textbf{46.3\%}.
Our acceleration method only causes a minor drop or consistent results in the segmentation performance.

The reduced token count leads to a \textbf{42.50\%} inference speedup and lowers GPU memory usage by \textbf{37.41\%} over SAM2. Averaging over all sequences from all datasets used in Table \ref{tab:main_results}, our method requires \textbf{2.6} simulated clicks per sequence compared to 4.2 clicks required in SAM2 baseline showing minimal human-in-the-loop requirement. Hence pruning the noisy information helps keep the $\mathcal{J\&F}$ scores consistently above a threshold of 80\%, falling below which click-prompts are leveraged.

%shows that our text-driven token pruning consistently speeds up over other SOTAs by 31.7\% while cutting GPU memory usage by 46.3\%, across all datasets. Any refinement is at $\mathcal{J\&F}<80\%$.

\begin{table*}[!htb]
\centering
\caption{\normalsize Comparison with SOTA methods across datasets.}
\resizebox{\textwidth}{!}{
\begin{tabular}{llcccccc}
\toprule
\textbf{Dataset} & \textbf{Method} & $\mathcal{J} \uparrow$ & $\mathcal{F} \uparrow$ & $\mathcal{J\&F} \uparrow$ & \textbf{FPS} $\uparrow$ & \textbf{VRAM (GB)} $\downarrow$ & \textbf{Refinements} \\
\midrule

\multirow{8}{*}{UVO~\cite{wang2021unidentifiedvideoobjectsbenchmark}}
& SAM2~\cite{ravi2024sam} & 0.614 & \cellcolor{third}0.987 & 0.805 & 17.9 & 2.35 & 4 \\
& STM~\cite{oh2019video} & 0.706 & \cellcolor{best}0.988 & \cellcolor{third}0.847 & 18.4 & \cellcolor{third}1.39 & -- \\
& AOT~\cite{yang2021associating} & \cellcolor{third}0.768 & 0.837 & 0.802 & 23.0 & 1.67 & -- \\
& DeAOT~\cite{yang2022decoupling} & 0.761 & 0.832 & 0.797 & 22.2 & 1.87 & -- \\
& XMem~\cite{cheng2022xmem} & 0.706 & \cellcolor{best}0.988 & \cellcolor{third}0.847 & \cellcolor{third}24.1 & 1.70 & -- \\
& Cutie~\cite{cheng2024putting} & 0.701 & 0.981 & 0.841 & 21.3 & 1.86 & -- \\
& \textbf{Ours} & \cellcolor{best}0.813 & 0.903 & \cellcolor{best}0.858 & \cellcolor{best}30.5 & \cellcolor{best}1.26 & \cellcolor{best}2 \\
\midrule

\multirow{8}{*}{PUMaVOS~\cite{Bekuzarov_2023_ICCV}}
& SAM2~\cite{ravi2024sam} & \cellcolor{best}0.891 & \cellcolor{third}0.985 & \cellcolor{best}0.938 & 17.6 & 2.27 & 4 \\
& STM~\cite{oh2019video} & 0.448 & 0.981 & 0.714 & 5.1 & 1.81 & -- \\
& AOT~\cite{yang2021associating} & 0.878 & 0.914 & 0.896 & 6.9 & 1.93 & -- \\
& DeAOT~\cite{yang2022decoupling} & \cellcolor{third}0.880 & 0.914 & 0.897 & 6.3 & 2.05 & -- \\
& XMem~\cite{cheng2022xmem} & 0.873 & 0.983 & 0.928 & \cellcolor{third}24.5 & \cellcolor{third}1.64 & -- \\
& Cutie~\cite{cheng2024putting} & 0.878 & 0.980 & \cellcolor{third}0.929 & 21.9 & 1.78 & -- \\
& \textbf{Ours} & 0.838 & \cellcolor{best}0.987 & 0.912 & \cellcolor{best}24.6 & \cellcolor{best}1.22 & \cellcolor{best}3 \\
\midrule

\multirow{8}{*}{EndoVis~\cite{allan2021stereo}}
& SAM2~\cite{ravi2024sam} & \cellcolor{best}0.894 & 0.981 & \cellcolor{best}0.938 & 16.2 & 8.90 & 5 \\
& STM~\cite{oh2019video} & 0.809 & \cellcolor{third}0.984 & 0.897 & 11.5 & 9.21 & -- \\
& AOT~\cite{yang2021associating} & 0.752 & 0.832 & 0.792 & 13.8 & 8.18 & -- \\
& DeAOT~\cite{yang2022decoupling} & 0.742 & 0.825 & 0.783 & 13.2 & 8.58 & -- \\
& XMem~\cite{cheng2022xmem} & \cellcolor{third}0.879 & 0.979 & \cellcolor{third}0.929 & 20.5 & \cellcolor{third}7.82 & -- \\
& Cutie~\cite{cheng2024putting} & \cellcolor{third}0.879 & 0.907 & 0.893 & \cellcolor{third}18.7 & 8.34 & -- \\
& \textbf{Ours} & 0.834 & \cellcolor{best}0.989 & 0.911 & \cellcolor{best}21.6 & \cellcolor{best}7.63 & \cellcolor{best}3 \\
\midrule

\multirow{4}{*}{VOST~\cite{tokmakov2023breaking}}
& SAM2~\cite{ravi2024sam} & 0.582 & 0.967 & 0.775 & 14.6 & 1.84 & 4 \\
& STM~\cite{oh2019video} & 0.327 & 0.976 & 0.651 & 9.25 & 7.41 & -- \\
& AOT~\cite{yang2021associating} & \cellcolor{third}0.822 & \cellcolor{best}0.983 & \cellcolor{third}0.902 & 15.71 & 9.31 & -- \\
& DeAOT~\cite{yang2022decoupling} & \cellcolor{best}0.942 & 0.959 & \cellcolor{best}0.951 & 16.03 & 10.42 & -- \\
& XMem~\cite{cheng2022xmem} & 0.674 & 0.970 & 0.822 & 18.5 & \cellcolor{third}1.52 & -- \\
& Cutie~\cite{cheng2024putting} & 0.686 & 0.977 & 0.831 & \cellcolor{third}18.7 & 8.34 & -- \\
& \textbf{Ours} & 0.719 & \cellcolor{third}0.981 & 0.850 & \cellcolor{best}22.3 & \cellcolor{best}1.18 & \cellcolor{best}3 \\
\midrule

\multirow{4}{*}{LVOSv2~\cite{hong2024lvos}}
& SAM2~\cite{ravi2024sam} & 0.625 & 0.969 & 0.797 & 12.8 & 2.01 & 4 \\
& STM~\cite{oh2019video} & 0.489 & \cellcolor{third}0.981 & 0.735 & 10.4 & 8.37 & -- \\
& AOT~\cite{yang2021associating} & 0.822 & \cellcolor{best}0.983 & 0.902 & 15.8 & 8.78 & -- \\
& DeAOT~\cite{yang2022decoupling} & \cellcolor{best}0.913 & 0.939 & \cellcolor{best}0.926 & 13.55 & 8.81 & -- \\
& XMem~\cite{cheng2022xmem} & 0.701 & 0.974 & 0.838 & 16.7 & \cellcolor{third}1.63 & -- \\
& Cutie~\cite{cheng2024putting} & \cellcolor{third}0.886 & 0.957 & \cellcolor{third}0.922 & \cellcolor{third}18.7 & 8.34 & -- \\
& \textbf{Ours} & 0.829 & \cellcolor{third}0.981 & 0.905 & \cellcolor{best}19.6 & \cellcolor{best}1.12 & \cellcolor{best}2 \\
\bottomrule
\end{tabular}
}

\label{tab:main_results}
\end{table*}

\subsection{Ablation Studies}
\subsubsection{Semantic and Uncertainty Contributions.}
In table~\ref{tab:ablation_signals}, The observed performance gains on the UVO dataset~\cite{wang2021unidentifiedvideoobjectsbenchmark} (averaged over all sequences) demonstrates that incorporating both semantic relevance and uncertainty estimation into the token pruning process leads to consistent improvements over using semantic cues alone, as visually ambiguous or occluded regions are no longer prematurely pruned.

\FloatBarrier
\FloatBarrier

\begin{table}[H]
\centering
\caption{\normalsize Signal-wise ablation on UVO.}
\label{tab:ablation_signals}
\begin{tabular}{lcc}
\toprule
\textbf{Signal Used} & $\mathcal{J\&F} \uparrow$ & \textbf{Retained Tokens} \\
\midrule
Text Prompt Only & 83.9 & 57 \\
\textbf{Text Prompt + Uncertainty} & \cellcolor{best}85.8 & 59 \\
\bottomrule
\end{tabular}
\end{table}

\subsubsection{ConvNeXt for Local Precision}
\begin{figure}[t]
  \centering
  \includegraphics[width=\linewidth]{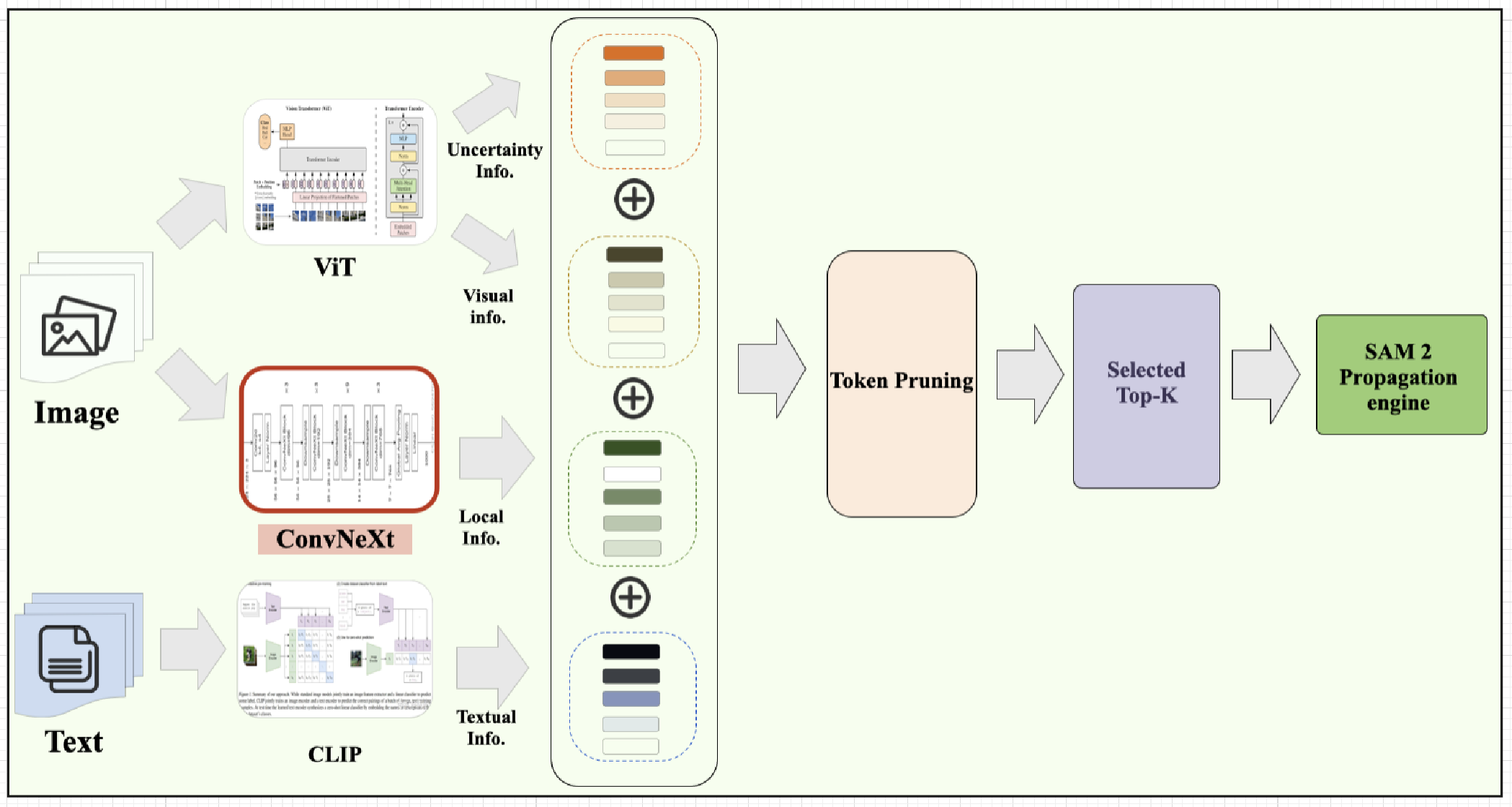}
  \caption{A frozen ConvNeXt~\cite{liu2022convnet2020s} backbone appended alongside the ViT encoder to capture fine-grained local patterns}
  \label{fig8}
\end{figure}

We append a frozen ConvNeXt~\cite{liu2022convnet2020s} backbone alongside the ViT encoder, as shown in Fig~\ref{fig8} to capture finer local spatial patterns, which might be ignored by ViT's which are more inclined on capturing the macroscopic features and their embeddings lack the finer parts below a certain scale, even with a hierarchical scale facility, as in ViT-Hiera. Table~\ref{tab:ablation_convnext} shows minor yet consistent improvements, on UVO~\cite{wang2021unidentifiedvideoobjectsbenchmark} (averaged over all sequences).

% \begin{figure}[t]
% \centering
% \includegraphics[width=0.47\textwidth]{AnonymousSubmission/LaTeX/ablation_1.png}
% \caption{A frozen ConvNeXt~\cite{liu2022convnet2020s} backbone appended alongside the ViT encoder to capture fine-grained local patterns}
% \label{fig:convnext}
% \end{figure}

\begin{table}[H]
\centering
\caption{\normalsize Effect of ConvNeXt integration (on UVO)}
\label{tab:ablation_convnext}
\begin{tabular}{lcc}
\toprule
\textbf{Visual Encoder} & $\mathcal{J\&F} \uparrow$ & \textbf{Overhead (ms)} \\
\midrule
ViT only & 85.8 & -- \\
ViT + ConvNeXt & \cellcolor{best}86.0 & +18.2 \\
\bottomrule
\end{tabular}

\end{table}

\subsection{Token Retention Sensitivity}
We evaluate pruning aggressiveness experiment on UVO~\cite{wang2021unidentifiedvideoobjectsbenchmark} (averaging over all sequences) by varying token retention rates. As shown in Table~\ref{tab:retention}, optimal balance is found near 30\%, which suppresses distractors while maintaining semantic coverage.

\begin{table}[H]
\centering
\caption{\normalsize Varying token retention and its effect.}
\label{tab:retention}
\begin{tabular}{ccc}
\toprule
\textbf{Retention (\%)} & $\mathcal{J\&F}$ ↑ & \textbf{FPS} ↑ \\
\midrule
100 (No pruning) & 85.6 & 17.9 \\
50 & 85.7 & 23.2 \\
30 & \cellcolor{best}85.8 & \cellcolor{best}30.5 \\
10 & 84.9 & 42.8 \\
\bottomrule
\end{tabular}

\end{table}

\subsection{Monte Carlo Pass Sensitivity}
We analyze the effect of the number of Monte Carlo Dropout passes $T$ used for uncertainty estimation on UVO~\cite{wang2021unidentifiedvideoobjectsbenchmark}, averaging results over all sequences. As shown in Table~\ref{tab:mcd}, increasing $T$ improves uncertainty stability and segmentation accuracy up to a point, after which additional passes introduce diminishing returns while incurring higher computational cost. We observe that $T=5$ offers the best trade-off between segmentation quality and inference efficiency, and therefore adopt it as the default setting.

\begin{table}[H]
\centering
\caption{\normalsize Effect of Monte Carlo passes $T$ on uncertainty estimation.}
\label{tab:mcd}
\begin{tabular}{ccc}
\toprule
\textbf{MC Passes ($T$)} & $\mathcal{J\&F}$ ↑ & \textbf{FPS} ↑ \\
\midrule
4 & 85.5 & \cellcolor{best}32.1 \\
5 & \cellcolor{best}85.8 & 30.5 \\
6 & 85.8 & 27.6 \\
\bottomrule
\end{tabular}
\end{table}

\subsection{Robustness to Automated and Human Prompts}
\label{sec:prompt_ablation}

We study the sensitivity of our pruning mechanism to different sources and qualities of textual prompts. In practical deployment, text prompts may be provided directly by a user or automatically generated (hence found vague at some random generation). Importantly, the text signal in our framework is not intended to perfectly encode user intent, but rather to act as a semantic prior that guides token selection. We therefore evaluate three representative prompt settings: (i) an accurate automated prompt generated from an object-centric region description, (ii) a vague automated prompt with some semantic specificity, and (iii) a human-provided prompt. As shown in Table~\ref{tab:prompt_ablation}, all three settings yield similar segmentation accuracy for the video sequence shown in fig.~\ref{fig1}, indicating that the pruning strategy remains stable even under noisy or under-specified semantic guidance.

\begin{table}[H]
\centering
\caption{\normalsize Effect of prompt specificity and source on segmentation accuracy and efficiency (UVO dataset, averaged over all sequences).}
\label{tab:prompt_ablation}
\begin{tabular}{lc}
\toprule
\textbf{Prompt Type} & $\mathcal{J\&F}$ ↑ \\
\midrule
(i) brown fluffy dog during gentle face grooming & \cellcolor{best}85.80 \\
(ii) puppy being handled & \cellcolor{third}85.78 \\
(iii) dog being groomed & \cellcolor{best}85.80 \\
\bottomrule
\end{tabular}
\end{table}

\section{Discussion and Future Work}
\label{sec5}
Our work presents a text-driven token pruning approach to accelerate SAM2 as the vision foundation model for video object segmentation.
We reduce visual tokens produced by the SAM2 vision encoder to only retain tokens relevant to the object being prompted.
We argue that eliminating redundant information at the beginning of the segmentation pipeline can prevent the model from carrying the computational burden of these tokens throughout the entire segmentation process.
This is reflected in our result in Table~\ref{tab:main_results}, which shows that our method consistently yields competitive acceleration on the inference speed against prior SOTA models.
Additionally, in terms of segmentation performance, our method only causes a minor drop on the baseline SAM2.
We argue that this capability is the effect of the elimination of the noisy information from redundant visual tokens at the beginning of the pipeline, as well as the restoration of a few ambiguous tokens through Monte Carlo dropout.
Moreover, as our method does not perform any modification to the VOS model, we argue that our method can be integrated into other methods that attempt to accelerate the inference speed of the SAM2 (e.g., EfficientTAM~\cite{xiong2024efficient, xiong2025efficient}).
We leave the exploration on this direction for future work.

\section{Conclusion}
\label{sec6}
We presented \textbf{Fast SAM2 with text-driven token pruning} technique applied post image encoder and pre-memory in SAM2, reducing compute in the propagation stack while preserving accuracy. By integrating semantic relevance, uncertainty, and visual context, we obtain substantial improvements in terms of inference speed and GPU memory efficiency with minimal architectural overhead. Our expanded evaluations and robustness studies demonstrate the viability of token pruning as a practical acceleration strategy for real-world video segmentation pipelines.

\section{Acknowledgements}
This work was supported by the National Natural Science Foundation of China (Grant No. 62572104). 
The first author conducted this research during a summer internship at the University of Electronic Science and Technology of China (UESTC), Chengdu, China.
% \appendix
% \section{Example Appendix Section}


% %% For citations use: 
% %%       \cite{<label>} ==> [1]

% %%
% Example citation, See \cite{lamport94}.

% %% If you have bib database file and want bibtex to generate the
% %% bibitems, please use
% %%
% %%  \bibliographystyle{elsarticle-num} 
% %%  \bibliography{<your bibdatabase>}

% %% else use the following coding to input the bibitems directly in the
% %% TeX file.

% %% Refer following link for more details about bibliography and citations.
% %% https://en.wikibooks.org/wiki/LaTeX/Bibliography_Management

\end{document}